\documentclass[aps,preprint,onecolumn,superscriptaddress,showpacs,showkeys,nofootinbib]{revtex4-1}

\usepackage{graphicx}
\usepackage{hyperref}
\usepackage{chngcntr}
\usepackage{subcaption}
\usepackage{amsmath}
\usepackage{amssymb}
\usepackage{lineno}
\usepackage{listings}
\usepackage{xcolor}

\definecolor{codegreen}{rgb}{0.0,0.6,0.0}
\definecolor{codeblue}{rgb}{0.0,0.0,0.8}
\definecolor{codepurple}{rgb}{0.5,0.0,0.5}
\definecolor{codered}{rgb}{0.8,0.0,0.0}
\definecolor{codegray}{rgb}{0.98,0.98,0.98}
\begin{document}

\title{A diversity-enhanced genetic algorithm for efficient exploration of parameter spaces}

\author{Jonas Wess\'en}\thanks{Equal contribution.}
\affiliation{University of Toronto, Department of Biochemistry, 1 King's College Circle, Toronto, Ontario, M5S 1A8, Canada}

\author{Eliel Camargo-Molina}\thanks{Equal contribution. Corresponding author: \href{mailto:eliel.camargo-molina@speldesign.uu.se}{eliel.camargo-molina@speldesign.uu.se}}
\affiliation{Uppsala University, Department of Physics and Astronomy / Department of Game Design, Box 516, SE-751 20, Uppsala / Visby, Sweden}

\begin{abstract}
We present a Python package together with a practical guide for the implementation of a lightweight diversity-enhanced genetic algorithm (GA) approach for the exploration of multi-dimensional parameter spaces. Searching a parameter space for regions with desirable properties, e.g.~compatibility with experimental data, poses a type of optimization problem wherein the focus lies on pinpointing all ``good enough'' solutions, rather than a single ``best solution''. Our approach dramatically outperforms random scans and other GA-based implementations in this aspect. We validate the effectiveness of our approach by applying it to a particle physics problem, showcasing its ability to identify promising parameter points in isolated, viable regions meeting experimental constraints. The companion Python package is applicable to optimization problems beyond those considered in this work, including scanning over discrete parameters (categories). A detailed guide for its usage is provided.
\end{abstract}

\maketitle

\section{Introduction}

The scientific method relies on the interplay between theory and experiment. From the theory side, mathematical models are constructed to describe the phenomena observed in nature. In general, these models have free parameters that need to be chosen before the model can be used to make predictions, which, for complex models, amounts to a challenging optimization problem. Numerical methods for parameter space scans are thus a cornerstone of progress in almost every scientific field. It enables the testing of theoretical models by driving the search of input parameters that make predictions in agreement with experimental results. 

For example, in the field of high-energy physics, the Standard Model (SM) of particle physics has been a remarkably successful theory, predicting a wide range of phenomena with high precision. However, the SM does not explain observed phenomena such as dark matter, dark energy or the fact that neutrinos are not massless. To address these shortcomings, physicists construct models that extend it. Such models, so-called beyond the SM (BSM), generally postulate the existence of new quantum fields (describing particles). Besides the same parameters as the SM, each new field adds new mass- and interaction parameters. All parameters need to be carefully chosen to not conflict with experimental results \cite{PDG2022}. In other words, for a new model to be viable, it needs to have regions in its parameter space that make correct predictions for all the observables that have been measured so far, while also offering answers to the open questions the SM leaves behind.

Locating a single workable set of parameters (a parameter point) is often of little use, and the goal is typically to find all the regions within the model's parameter space that harbour viable parameter points. Finding diverse sets of such points is challenging when dealing with parameter spaces of high dimensionality and when testing a single parameter point (calculating all of its predictions) implies performing computationally demanding calculations of many observables. A simple ``random scan'', wherein uncorrelated random parameter values are first generated from pre-defined probability distributions and then tested, is attractive for its ease of implementation but is severely constrained by the assumptions in the chosen probability distributions. Furthermore, it performs poorly when applied to parameter spaces with small viable regions that are difficult to find. Alternatively, there are various sophisticated tools \cite{de2023exploring} that use advanced statistical methods for BSM parameter space scans. These tools work well, though they also bring a steep learning curve and are difficult to customize.

A promising approach to this challenge is the use of genetic algorithms (GAs). GAs are are population-based optimization algorithms functioning through principles inspired by biological evolution \cite{holland1992adaptation,shapiro1999genetic,goldberg2007genetic}. In close analogy to biological evolution, each parameter point is an individual, and each of its parameters a gene. In GA parameter space searches, a population of parameter points co-evolve through offspring generation and selection towards parameter space regions exhibiting high ``fitness'', i.e.~regions compatible with the imposed constraints. The ease-of-implementation make GAs an attractive alternative, while dramatically outperforming random scans and in some cases matching more sophisticated methods. 

The power of GAs has been acknowledged in multiple disciplines, such as electronic engineering \cite{Vellasco:2001:EEA:559960}, climate modelling \cite{Stanislawska:2012:MGT:2400749.2401077}, computational quantum mechanics \cite{Sugawara2001}, large-language models \cite{Guo2023}, protein folding \cite{Unger1993, Pedersen1997}, protein phase-separation \cite{Chew2023}, and high-energy physics \cite{Allanach2004,Cranmer2005,Teodorescu2008,Akrami2010,Ruehle:2017mzq,Camargo-Molina:2017klw, luo2020genetic, biekotter2021reconciling}.

We have previously designed and implemented a GA to scan the parameter space of a Two-Higgs Doublet Model (2HDM), a simple extension of the SM with an extra Higgs boson, as a viability test of the method. This initial application served as a test case, allowing us to fine-tune our GA and assess its performance. The GA featured a novel diversity-enhanced selection procedure which prevented the algorithm from getting trapped at local fitness peaks and produced rich sets of viable yet different parameter points. Encouraged by these results, we next applied the same GA to find combinations of parameters in a Three-Higgs Doublet Model (3HDM) that would lead to visible new physics signals at the Large Hadron Collider. The results of this application, though crucially not the code of our GA, were published in \cite{Camargo-Molina:2017klw}. 

Recently, one of us applied a similar GA to study a space of amino-acid sequences, representing intrinsically disordered proteins, with respect to commonly used parameters that quantify the pattern of electric charge along the protein chain backbones \cite{Pal2024}. The charge-pattern is important for biological function as it impacts both the single-chain conformational ensemble \cite{Das2013,Sawle2015} and the protein's tendency to scaffold phase-separated droplets known as membrane-less organelles \cite{Lin2016,McCarty2019,Pal2021,Wessen2021,Wessen2022JCP,Wessen2022JPCB,Lin2023}. This GA application presented yet another example in which the diversity-enhanced selection provided a substantial increase in performance. 

Motivated by this and by a general renewed interest in GAs within particle physics phenomenology, we decided to revisit our original implementation. We have rewritten it as a Python module, making it readily installable via pip and with additional documentation. The source code and documentation can be found at:
\begin{center}
\href{https://github.com/JoseEliel/lightweight_genetic_algorithm}{github.com/JoseEliel/lightweight\_genetic\_algorithm} (Source Code)\\
\href{https://lightweight-genetic-algorithm.readthedocs.io/}{lightweight-genetic-algorithm.readthedocs.io} (Documentation)
\end{center}
Particular efforts have been made to make our GA accessible, versatile and user-friendly. Several impressive open-source Python libraries already exist for GA optimization, such as \texttt{pymoo} \cite{pymoo} and \texttt{PyGAD} \cite{gad2021pygad}, and we encourage the reader to turn to those for a comprehensive implementation of GAs. With our code, we instead want to provide a minimal and transparent GA using the diversity-enhanced selection which nevertheless can be applied to generic optimization problems beyond those considered in this work.

This article is structured as follows. In Section \ref{sec:overview}, we overview GAs as applied to parameter space scans and describe key ingredients such as the fitness function, mutations, crossover methods and our novel diversity-enhanced survivor selection procedure. In Section \ref{sec:performance}, we apply our GA to find experimentally viable points in a type-I 2HDM parameter space. Here, we demonstrate how the algorithm's performance depend on population size and crossover method, and how it performs compared to the random scan approach. Conclusions, including suggested guidelines for GA parameter scans, are given in Section \ref{sec:conclusions}. In \ref{sec:lga_documentation}, we provide documentation for our Python module \texttt{lightweight{\_}genetic{\_}algorithm}, covering installation, usage examples and a list of features included in the package. 

\section{Genetic algorithms for parameter space scans} \label{sec:overview}
\subsection{Algorithm overview}

\begin{figure}
\centering  
\includegraphics[width=0.4\textwidth]{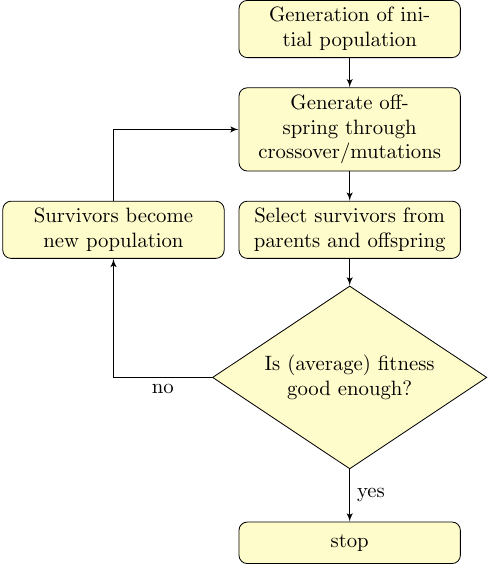}%
\caption{Flowchart of the main algorithm. Beginning with a randomly generated population, the GA proceeds through iterative offspring generation and survivor selection until a satisfactory population of individuals has been reached.}
\label{fig:flowchart}
\end{figure}

The basic structure we will use in this work for GAs is shown in Fig.~\ref{fig:flowchart}. The goal of the algorithm is to produce a population $P$ of $n$ individuals $\lbrace I_j \rbrace_{j=1}^n$ (parameter points), where each individual $I_j$ is a multi-component object with components $(I_j)_k$ (genes) representing the coordinates (parameters) of a point in parameter space. Here, the index $k=1,\dots,N$ runs over the components of $I_j$ and $N$ is the dimensionality of the parameter space. The GA is deemed successful if $P$ contains individuals that are both sufficiently spread out and have a high fitness, measured with the fitness function $f(I_j)$.

As shown in Fig.~\ref{fig:flowchart}, the GA begins by randomly generating a population of $n$ individuals. Next, the individuals undergo reproduction to generate a new set of offspring individuals based on the parent individuals' parameters. A subset of $n$ survivor individuals are selected out of the combined set of parent and offspring individuals to form the next generation. The process repeats until a satisfactory population is found or a predetermined number of generations have been completed. 

The fitness function, the offspring creation method and the survivor selection procedure defines the specific GA implementation. In the following, we describe the details of our light-weight implementation.

\subsection{Initial population}

The initial population consists of $n$ individuals. We generate the initial population using a uniform random distribution for each parameter within a specified range. This input range can be determined by theoretical constraints, such as perturbativity or unitarity, or it may simply define a preferred area of study. 

\subsection{Fitness function} \label{sec:fitness}

The fitness function is a key component of the genetic algorithm (GA), as it evaluates and compares the performance of individuals and helps determine their survival. It is a mapping $f: I_j  \rightarrow \mathbb{R}$, where $I_j$ represents an individual. If $I_j$ provides a better solution to the problem at hand than $I_k$, then $f(I_j) > f(I_k)$.

The fitness function for BSM theory parameter space scans quantifies to what degree a parameter point satisfies experimental and theoretical constrains. It is very useful to categorize these into two groups: hard constraints (HCs) and soft constraints (SCs). HCs are conditions that an individual must satisfy to be considered a valid solution. If an individual $I_j$ violates any HC, it is immediately deemed irrelevant and is attributed a very small fitness. If the individual $I_j$ satisfies all HCs, the algorithm proceeds to evaluate the SCs. SCs are conditions that reflect, for example, the statistical likelihood that a physical observable $O(I_j)$ is in agreement with a noisy experimental measurement of $O$.

In many cases, evaluating SCs is more computationally demanding than evaluating HCs. Therefore, the efficiency of the GA is enhanced by this two-step fitness construction, as the SCs are evaluated only for the subset of individuals $I_j$ in the population that satisfy all HCs. 

In practice, the relative weights between different contributions to the fitness determine the order in which the GA focuses on optimizing the associated constraints. Earlier generations tend to focus on the HCs, while the SCs are optimized mainly during later generations. To reduce the number of generations focusing on HCs, it is crucial that their associated fitness contributions can guide the population towards regions in parameter space where the HCs are satisfied.

\subsection{Reproduction: Crossover and Mutation} \label{sec:crossover_methods}

Reproduction in GAs involves two main processes: crossover and mutation. Crossover describes how new individuals (offspring) are generated by combining the properties of two parent individuals. Mutation, on the other hand, introduces random changes in the offspring, which ultimately drives the exploration of parameter space regions beyond those of the initial population.

There are in principle many ways to implement crossover, both when it comes to the method with which new parameters are generated but also in deciding how to pair the parents. In this work we have selected a number of representative options that we will benchmark in the Section \ref{sec:performance}.

\subsubsection{Crossover methods}

Crossover is defined as a function $M: I_i \times I_j \rightarrow I_k$, where $I_i$ and $I_j$ are parents and $I_k$ are the offspring. Here, we discuss the four methods implemented in our algorithm: midpoint crossover, either/or crossover, between crossover, and no crossover.

\begin{itemize}
    \item \textbf{Midpoint Crossover:} This method generates offspring by taking the arithmetic average of the parent individuals' parameters, i.e., $(M[I_i, I_j])_k = [(I_i)_k + (I_j)_k]/2$.
    
    \item \textbf{Either/Or Crossover:} In this method, each parameter of the offspring is randomly selected from one of the parents. Formally, $(M[I_i, I_j])_k = (I_j)_k$ if $R_k  < 1/2$, or $(I_i)_k$ if $R_k  > 1/2$, where $R_k$ is a uniform random number in the interval $[0,1]$.
    
    \item \textbf{Between Crossover:} This method generates offspring's genes (parameters) at random points along the line segment between the parents' genes. Formally, $(M[I_i, I_j])_k = R_k$, where $R_k$ is a uniform number in the interval $[\min((I_i)_k, (I_j)_k), \max((I_i)_k, (I_j)_k)]$.
    
    \item \textbf{No Crossover:} In this method, offspring are generated as mutations of single parents, without any crossover.
\end{itemize}

These methods are implemented in our Python package, documented in \ref{sec:lga_documentation}.

\subsubsection{Pairing methods}

Aside from the crossover method, one also needs to decide which individuals get to participate in crossover. We consider the two following extremal scenarios:
\begin{itemize}
    \item \textbf{$\mathcal{O}(n^2)$:} This method performs crossover for every pair of individuals in the population, resulting in $n(n-1)/2$ offspring per generation.
    \item \textbf{$\mathcal{O}(n)$:} This method selects $n$ random pairs for crossover, yielding $n$ offspring per generation.
\end{itemize}

\subsubsection{Mutation}

Mutation introduces random changes in the offspring's parameters which is important for the GA to not be constrained to the parameter space region populated by the initial population. Once an offspring individual $I_i$ has been produced through one of the above described crossover methods, each of it's genes $(I_i)_k$ has a probability $p_{\rm mutation}$ of being mutated. For numerical genes, we consider \textit{additive} and \textit{multiplicative} mutations, which imply $(I_i)_k \rightarrow R + (I_i)_k$ and $(I_i)_k \rightarrow R \cdot (I_i)_k$, respectively, where $R$ is a random number. For additive mutations, $R$ is drawn from a normal distribution with zero mean and standard deviation $\sigma = \left( g_k^{\rm (max)} - g_k^{\rm (min)} \right)/10$ where $[g_k^{\rm (min)}, g_k^{\rm (max)}]$ is the initial range for gene $k$ used to generate the initial population. For multiplicative mutations, $R$ is a normally distributed with mean 1 and standard deviation 0.5. Given that a numerical gene has been selected for mutation, the mutation mode is selected at random with equal probability (50\%) for each mode. Categorical genes are instead mutated by replacing $(I_i)_k$ by a new gene value selected at random (with uniform probability) from the given allowed categories. By default we choose $p_{\rm mutation} = 0.1$.

\subsection{Diversity-enhanced survivor selection} \label{sec:diversity_enhanced_selection}

In genetic algorithms, the process of survivor selection needs to ensure the preservation of the best individuals across generations while simultaneously encouraging the algorithm to explore new parameter space regions. In this work, we use a common strategy, known as ``elitism'', wherein the survivors are selected out of the combined set of parents and offspring individuals. This helps ensure that the most fit individuals are kept and that the quality of solutions does not degrade over time.

However, while elitism helps to maintain the quality of solutions, it may lead to a lack of diversity in the population over time, i.e. after sufficiently many generations all individuals in the population will be effectively clones of each other. To counteract this, we introduce a method that combines fitness with a penalty for similarity among individuals, thereby promoting diversity in the population.

In this method, the survivors are iteratively selected according to best fitness. However, after each survivor is selected, a penalty $D(I_k,I_s)$ is subtracted from the fitness of all remaining individuals $I_k$, i.e.
\begin{equation}
f(I_k) \rightarrow f(I_k) - D(I_k,I_s). 
\end{equation}
This penalty is substantial if $I_k$ and $I_s$ are similar, thereby discouraging the survival of similar individuals and promoting diversity.

The penalty function $D(I_j,I_k)$ is defined as
\begin{equation}\label{eq:diversity_penalty}
D(I_j,I_k) = D_0 \, \mathrm{e}^{-r^2(I_j,I_k)/r^2_0} , 
\end{equation}
where the function $r(I_j,I_k)$ represents the ``distance'' between individuals $j$ and $k$, and $D_0$ is the penalty for exactly overlapping individuals, $D(I_j,I_j)=D_0$. The parameter $r_0$ defines the characteristic distance below which individuals are considered similar and consequently get non-negligible penalties during selection. The diversity penalties are reset between generations such that each generation's selection is initiated with un-penalized fitnesses.

The most appropriate choice of distance function generally depends on the specifics of the application. A traditional Euclidean distance measure, 
\begin{equation} \label{eq:euclidean_distance_measure}
r^2(I_j,I_k) = \sum_{i=1}^N \left[ (I_j)_i - (I_k)_i \right]^2 \,\,\, \mbox{(Euclidean)}
\end{equation}
is a natural starting point, but it may not be the most suitable for parameter space scans. This is because it does not account for the scale differences among parameters. For instance, if one parameter typically ranges from 0 to 1, while another ranges from 0 to 1000, the Euclidean distance would be dominated by the larger-scale parameter. This could skew the results and lead to a less effective exploration of the parameter space.

For this reason, we have found that the following dynamically weighted Euclidean distance function,
\begin{equation} \label{eq:dynamic_distance_measure}
r^2(I_j,I_k) = \sum_{i=1}^N \frac{ \left[ (I_j)_i-(I_k)_i \right]^2}{\left( |(I_j)_i|+|(I_k)_i| + \epsilon \right)^2} \,\,\, \mbox{(Dynamic)}
\end{equation}
to work well for the type of parameter scan considered in Section \ref{sec:performance} and in \cite{Camargo-Molina:2017klw}. Here, $\epsilon$ is a small regulator added to avoid zeros in the denominators. In our Python package, documented in \ref{sec:lga_documentation}, it is possible to choose either the Euclidean distance or the dynamic measure defined above. It is also possible for the user to implement their own distance measure.

By default in our Python implementation, we set $D_0 = 1$ and set $r_0$ based on the initial population. Specifically, we compute the average squared distance between all distinct pairs of individuals in the initial population:
\begin{equation}
\langle r^2 \rangle = \frac{1}{N'} \sum_{i \neq j} r^2(I_i, I_j),
\end{equation}
where $N'$ is the number of unique pairs $(i, j)$ with $i \neq j$. Then, we define:
\begin{equation}
r_0 = \frac{ \sqrt{ \langle r^2 \rangle } }{10}.
\end{equation}
In other words, we set the characteristic distance $r_0$ to be one-tenth of the typical separation between individuals in the initial population.

\subsection{Adaptations for categorical genes}\label{sec:adaptations_categorical}

While the standard genetic algorithm set-up specifies methods for numerical genes, it is straightforward to adapt these methods for categorical genes as well. Categorical genes represent discrete, non-numeric options, such as categories or labels (e.g.~the labels defining a protein's amino-acid sequence).

To adapt the genetic algorithm for categorical genes, minor changes are required in how crossover and mutation operations are performed:

\begin{itemize} \item \textbf{Crossover}: An effective method for categorical genes is the ``Either Or'' method. This method maintains the integrity of discrete categories and avoids the complications of averaging or interpolating non-numeric values.

\item \textbf{Mutation}: For mutation, let $g_i$ represent a categorical gene chosen from a set $C$. The simplest option is that each gene $ g_i $ is randomly swapped with a new category $g_j \in C$ from its possible set, rather than modifying a numeric value.

\item \textbf{Distance Measure}: The diversity enhancement works here much in the same way, though with a corresponding categorical measure. A good measure to use is the Hamming distance, which calculates the proportion of differing elements between two categorical strings. Formally, for two categorical lists $ x $ and $ y $ of length $ n $, the Hamming distance $r^2_{\rm H}$ is defined as:
\begin{equation} \label{eq:hamming_distance_measure}
r^2_{\rm H}(x, y) = \frac{1}{N} \sum_{i=1}^{N} (1-\delta_{x_i y_i})
\end{equation}
where $\delta_{ab}$ is 1 if $a = b$ and 0 otherwise. This measure effectively ensures proper diversity without misconstruing categorical disparities as numerical distances. By default we set both $D_0$ and $r_0$ to 1 for categorical genes.

\end{itemize}

\section{Assessing the performance} \label{sec:performance}

In this section, we assess the performance of the algorithm outlined in the previous section and provide guidelines for the implementation of GAs in parameter space scans.

To assess the GA, we performed a series of benchmarking studies based on scanning the parameter space of a type-I Two-Higgs Doublet Model (2HDM) \cite{Branco:2011iw}. The 2HDM extends the Standard Model of particle physics by postulating the existence of an additional Higgs doublet field. This 2HDM introduces seven new parameters beyond the SM that need to be chosen in a way consistent with all experimental contraints. Given a set of parameters, the predictions for observables can be calculated with the help of established open-source tools and their output can be used to build the fitness function. Below we describe this in more technical detail.

The model has seven free parameters: $m_h$, $m_H$, $m_A$, $m_{H^{\pm}}$, $\sin(\beta - \alpha)$, $m_{12}^2$, and $\tan \beta$. These parameters correspond to the masses of the new BSM particles, i.e. two neutral CP-even states, the mass of the neutral CP-odd state, the mass of the charged Higgs boson, and the scalar-pseudoscalar mixing angle parameter, the soft $\mathbb{Z}_2$ symmetry breaking parameter, and the ratio of the vacuum expectation values of the two Higgs doublets, respectively. 

The purpose of the parameter scan is to identify the regions in the 2HDM parameter space that evade tight experimental bounds on BSM physics. The fitness function used to evaluate the performance of the GA is constructed using a combination of established phenomenology tools, each serving a specific purpose. We used HiggsBounds 4.3.1 \cite{HiggsBounds} to cross-check the Higgs sector bounds against the experimental cross-section limits from the Tevatron, LEP, and LHC. We used 2HDMC 1.7.0 \cite{2HDMC} to evaluate the properties of the potential, determine Higgs masses and couplings using tree-level relations, and calculate decay widths at the leading order. Finally, we used HiggsSignals 1.4.0 \cite{HiggsSignals} as it performs a statistical test of the Higgs sector predictions for a given model, in this case the 2HDM.\footnote{We note here that our benchmarking studies were performed back in 2016, and as such use the latest versions of the codes available at the time.}

The initial population was chosen randomly within predefined intervals:

\begin{minipage}[t]{0.45\textwidth}
  \begin{align*}
  10 & \leq m_h \leq 125 \, \text{GeV} \\
  126 & \leq m_H \leq 500 \, \text{GeV} \\
  10 & \leq m_A \leq 500 \, \text{GeV} \\
  90 & \leq m_{H^{\pm}} \leq 500 \, \text{GeV} \\
  \end{align*}
\end{minipage}%
\begin{minipage}[t]{0.45\textwidth}
  \begin{align*}
  -1 & \leq \sin(\beta - \alpha) \leq 1 \\
  0 & \leq m_{12}^2 \leq 1000 \, \text{GeV}^2 \\
  2 & \leq \tan \beta \leq 25 \\
  \end{align*}
\end{minipage}

The fitness function is constructed as follows:

{\bf Hard Constraints (HCs)}
\begin{itemize}
\item 2HDMC checks for the stability of the potential and the tree-level unitarity of the scattering matrix. If the potential is unstable or the scattering matrix is non-unitary, the fitness is set to $-100$.
\item If the point violates the perturbativity of the quartic couplings, the fitness is set to $-100$.
\item If a 2HDMC fails to calculate $S^\prime$ and $T^\prime$ parameters, a fitness of $-100$ is added as punishment.
\item If $m_A + m_h <  m_Z$, a fitness value of $-100 - m_Z + m_A + m_h$ is assigned guaranteeing the consistency with $Z$-width measurements. 
\item When $ m_{H^{\pm}} < 80\, \text{GeV}$, the point gets the fitness of $-100 - (80- m_{H^{\pm}})$.  
\item Any scalar mass $m$ smaller than 10 GeV results in assignment of $-100-(10-m)$ as fitness to that point. 
\end{itemize}

{\bf Soft Constraints (SCs)}
\begin{itemize}
\item If the $S^\prime$ and $T^\prime$ parameters, calculated by 2HDMC, fall outside the 90\% confidence level (CL) region, which is represented by an ellipse, a penalty is applied. The penalty is calculated as $-10 - E(S,T)$, where $E$ represents the fitted ellipse equation. This penalty is larger for points that are further from the ellipse, with the value of 10 chosen to ensure the penalty is of the same order of magnitude as other contributions.
\item For each channel that HiggsBounds flag as excluding the parameter point, a penalty of $-10$ contributes to the fitness. 
\item The fitness also includes the $p$-value calculated from HiggsSignals, which is added as  $\log_{10} p$. The $p$-value from HiggsSignals, derived from the $\chi$-squared statistic, assesses the probability that the experimental data align with Standard Model Higgs predictions. Lower $p$-values, emerging from higher $\chi$-squared values, suggest notable deviations indicative of new physics.
\end{itemize}

We assess the performance of the algorithm by monitoring the arithmetic average fitness $f_{\rm s}$ among survivors (i.e.~the individuals passing the survivor selection step) during the course of the parameter scan. A well-performing GA reaches a high $f_{\rm s}$ in a short amount of time producing a final population with a diverse set of physically viable parameter points. We use the number of fitness evaluations, rather than the number of generations, as a measure of runtime as the fitness evaluation is the most computationally expensive step of the algorithm in our set-up. This allows for comparing runs with different population sizes and number of survivors. 

Our fitness function is constructed such that if an individual $I_j$ satisfies all HCs, it has a fitness $f(I_j) > -1$ that coincides with $\log_{10} p$ where $p$ is the $p$-value from HiggsSignals. Is it therefore instructive to consider $\tilde{p} = 10^{f_{\rm s}}$, rather than $f_{\rm s}$, to showcase the quality of the parameter points in the population. During the later stages of the algorithm, $\tilde{p}$ corresponds to the geometric average of the HiggsSignals $p$-values of the surviving individuals. 

All 2HDM GA parameter scans use diversity-enhanced selection with the dynamic distance measure in Eq.~\eqref{eq:dynamic_distance_measure}, setting $\epsilon=10^{-15}$, with fixed $D_0=1$ and $r_0=1$. 

\subsection{Population size}
To assess the impact of population size, we performed two separate runs with distinct implementations of a GA. Both runs began with an initial population of 350 individuals (parameter points), employing the ``Between'' crossover method and selecting $n=350$ survivors per generation. The ``$\mathcal{O}(n^2)$ offspring'' run performed crossover for every survivor pair, resulting in $n(n-1)/2=61075$ offspring for $n=350$. The ``$\mathcal{O}(n)$ offspring'' run, in contrast, selected $n$ random pairs for crossover, yielding only $n=350$ offspring per generation.

\begin{figure}[htbp]
  \centering
  \includegraphics{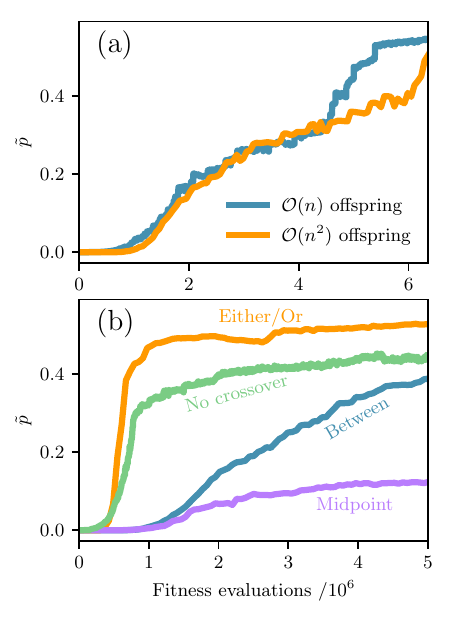}
  \caption{Fitness evaluations vs.~average $\tilde{p}$-value for all benchmark runs. The $\tilde{p}$-value is defined as $10^{f_{\rm s}}$ where $f_{\rm s}$ is the average fitness among selected survivors. For the later stages of the run it is the same as the $p$-value reported by HiggsSignals. (a) ``$\mathcal{O}(n^2)$ offspring'' and ``$\mathcal{O}(n)$ offspring''. (b) All crossover methods. Each curve shows $\tilde{p}$ averaged over 3 independent runs.}
  \label{fig:options_comparison}
\end{figure}

The evolution of $\tilde{p}$ for both runs is shown in Fig.~\ref{fig:options_comparison}a. We want to note that because we use number of fitness evaluations as the measure of runtime and not number of generations, the ``$\mathcal{O}(n)$ offspring'' run has many more generations than the ``$\mathcal{O}(n^2)$ offspring'' run. The figure suggests a similar performance between the two approaches despite the large difference in number of offspring per generation. However, the ``$\mathcal{O}(n)$ offspring'' run seem to perform slightly better towards the later stages of the run. We therefore conclude that the quality of the parameter scan is relatively robust against the choice of the number of offspring per generation, although the fine-tuning of  parameter points towards the end stages of the run likely benefit from the frequent survivor selections in the ``$\mathcal{O}(n)$ offspring'' approach.

\subsection{Crossover method}
Next, we investigate the impact on performance from the crossover and pairing methods outlined in Section \ref{sec:crossover_methods}. We performed three independent runs for each crossover method, all using random initial populations of size $n=350$. For the Midpoint-, Either/or- and Between crossover functions, we produce one offspring per $n=350$ randomly selected pairs of individuals in each generation, as in the ``$\mathcal{O}(n)$ offspring'' run of Fig.~\ref{fig:options_comparison}a. In the runs without crossover (No crossover), we generate one offspring per individual in the parent generation which also results in $n=350$ offspring per generation. The evolution of $\tilde{p}$, averaged over the three runs for each crossover method, is shown in Fig.~\ref{fig:options_comparison}b. This figure reveals that the choice of crossover method greatly impacts the efficiency of the GA to find parameter points with high fitness. The Either/or crossover method performs significantly better than both the Between- and Midpoint methods, the latter performing the worst out of all approaches considered. The runs without crossover perform slightly worse than the Either/or runs, but better than the runs with Between- and Midpoint crossover. We note that two out of the three independent scans without crossover performed just as good as the Either/or runs, while the third run ``lagged behind'' thus decreasing the average $\tilde{p}$ shown in Fig.~\ref{fig:options_comparison}b. It follows that, while a well-chosen crossover method has a positive impact on performance, a poorly chosen crossover method does more damage than good in comparison to the approach without crossover. We expect that the optimal crossover method can depend on the problem at hand. In contrast, the approach without crossover is both simple to implement and offers good performance, and is therefore an appealing method to test first. 

\subsection{Comparison with uniform sampling}

\begin{figure}[htbp]
  \centering
  \includegraphics{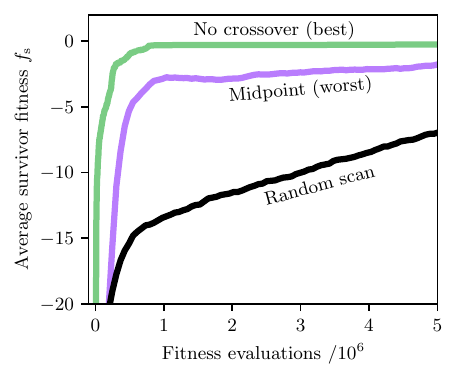}
  \caption{Average survivor fitness vs.~number of fitness evaluations for the best performing (no crossover, green curve) and worst performing (midpoint crossover, purple curve) GA setups compared with the same for a random scan (black curve). The GA implementations dramatically outperform the random scan in terms of quickly finding high-fitness parameter points (either considering the population average or the best individuals in the population).}
  \label{fig:RandomComp}
\end{figure}

Having demonstrated the impact of the population size and crossover method on the GA's performance, we next compare the GA to a random scan, i.e.~a parameter scan performed by generating uncorrelated parameter points according to a pre-defined probability distribution over the parameter space. To facilitate a comparison with our GA scans of the type-I 2HDM parameter space, we perform the random scan by generating parameter points from uniform distributions in the same input ranges as the GAs and compare the average survivor fitness $f_{\rm s}$ of the GA with the average fitness of the $n=350$ best randomly generated parameter points. In Fig.~\ref{fig:RandomComp}, we show the evolution of the average fitness of the random scan compared to the best and the worst performing GA scans of Fig.~\ref{fig:options_comparison}b. The worst performing GA scan is one of the Midpoint crossover runs and the best performing scan is one of the GA scans without crossover (performing near-identical to all three Either/or runs). Fig.~\ref{fig:RandomComp} shows that even the worst-performing GA parameter scan dramatically outperforms the random scan. We want to stress here that the complexity of the implementation of the random scan is comparable to that of the GA scans, as the random scan also requires the evaluation of the fitness function for each parameter point, thus the GA is by all considerations a preferable method.

\subsection{Comparison with SciPy's Differential Evolution}

To evaluate the exploration capabilities of our diversity-enhanced GA, we compared it against SciPy's Differential Evolution (DE). DE is an evolutionary algorithm similar to Genetic Algorithms, employing population-based search with operations like mutation, crossover, and selection \cite{5601760}. Since DE is widely used today, particularly through SciPy's implementation, it provides a relevant benchmark for comparison with our GA. Both algorithms were set with equivalent parameters: a population size of 200, running for 100 generations, and the same two-dimensional search space (with coordinates $x_1, x_2$). All other parameters where set as their default values. Our fitness function for comparison was:

\begin{equation} \label{eq:2D_fitness_function}
  f(x_1, x_2) = 
  \begin{cases} 
    -1000 & \text{if } |x_{1,2}| > 1.5 \\ 
    10 \cos(20 x_1 x_2) & \text{otherwise.} 
  \end{cases} 
\end{equation}

This function, featuring multiple global maxima, thoroughly tests the capability of the algorithms to find diverse solutions. We enforced the hard constraint $|x_i| \leq 1.5$ ($i=1,2$) to focus on a specific range of parameters. The factor 10 was chosen to ensure the fitness values and the penalty function (Eq.~\ref{eq:diversity_penalty}) are roughly of the same order of magnitude with default values.

\begin{figure}
  \centering
  \includegraphics[width=0.7\textwidth]{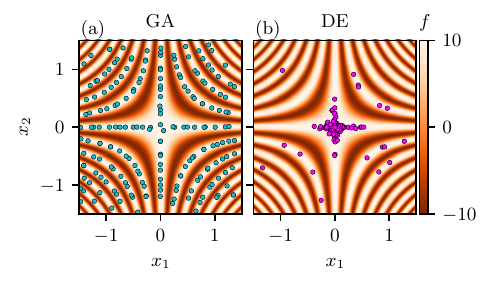}
  \caption{Comparison between (a) the GA and (b) SciPy's DE when used for locating maxima in a two-dimensional fitness landscape. The final populations are shown as scatter points and the fitness $f$, defined in Eq.~\eqref{eq:2D_fitness_function}, is shown as a heatmap. DE's final population is more concentrated around (0,0), while the GA explored the parameter space more effectively, finding a wider variety of good solutions.} 
  \label{fig:lga_vs_scipy} 
\end{figure}

In Fig.~\ref{fig:lga_vs_scipy}, we show the final population of one GA run (Fig.~\ref{fig:lga_vs_scipy}a) and one DE run (Fig.~\ref{fig:lga_vs_scipy}b). Our GA explored the entire function landscape more effectively, uncovering a wider array of global maxima with equally high fitness. DE, while significantly faster, centred around $x_1=x_2=0$ and found some of the function's maxima along many non-optimal solutions. Both algorithms' results demonstrate their respective efficiencies. However, the GA showed a superior exploratory performance with multiple high-fitness solutions spread across the parameter space.

Over 10 runs, the GA achieved an average fitness of \(9.91 \pm 0.02\) with an average spread of \(1.53 \pm 0.02\), while DE reached an average fitness of \(9.97 \pm 0.01\) with an average spread of \(0.5 \pm 0.2\). Here, the \textit{spread} refers to the average pairwise Euclidean distance between individuals in the final population, providing a measure of the diversity of solutions found by each algorithm. 

\section{Conclusions} \label{sec:conclusions}

In this paper, we have introduced a genetic algorithm for efficiently scanning high-dimensional parameter spaces for regions compatible with imposed constraints, as encapsulated by a fitness function. A key component of our algorithm is a selection procedure that promotes diversity through iterative fitness punishments of individuals similar to selected survivors. This causes the algorithm to broadly explore the parameter space for viable solutions, rather than converging on a single most optimal region. 

We have demonstrated the effectiveness of our genetic algorithm through a detailed benchmarking analysis focused on a parameter-scan task in particle physics, namely a scan of the type-I two Higgs doublet model. This analysis revealed that the efficiency is robust against the number of offspring per generation and that the choice of crossover method has a significant impact. In comparison to a mutation-only approach, a carefully chosen crossover method can yield small improvement in efficiency. However, a bad choice of crossover method can lead to a significant decrease in efficiency and we therefore expect that the mutation-only approach will be the most reliable choice for most applications. A comparison with a random scan revealed a dramatic improvement in efficiency, with the GA being able to reach average fitness values orders of magnitude better than the random scan, irrespective of the choice of crossover method.

Next, we compared our GA to the SciPy's differential evolution (DE) algorithm, which is a popular choice for parameter space scans. While the DE is faster than our GA, the GA is superior in terms of diversity showcasing the algorithms' respective strengths. We expect that the diversity aspect of our GA will be particularly important when exploring high-dimensional parameter spaces with multiple isolated regions harboring viable solutions. 

Finally, we have packaged our GA into a Python module \texttt{lightweight-genetic-algorithm}, which is available on GitHub and installable via pip. Installation instructions, features overview, user guide and examples are provided in \ref{sec:lga_documentation}. The module is versatile and well-documented, and we believe that it will be a valuable tool for researchers working on a wide range of problems that require the exploration of high-dimensional parameter spaces.

\section*{Acknowledgements}
The authors thank Rikard Enberg for his comments on the manuscript, Adriaan Merlevede for useful discussions regarding the use of GAs in computational biophysics and Tim Stefaniak for insightful discussions on 2HDMs and their phenomenology. JW acknowledges the generous financial support from Prof.~Hue Sun Chan at the University of Toronto. JW and ECM contributed equally to this work.

\bibliographystyle{apsrev4-1} 
\bibliography{bib}

\begin{thebibliography}{37}%
\makeatletter
\providecommand \@ifxundefined [1]{%
 \@ifx{#1\undefined}
}%
\providecommand \@ifnum [1]{%
 \ifnum #1\expandafter \@firstoftwo
 \else \expandafter \@secondoftwo
 \fi
}%
\providecommand \@ifx [1]{%
 \ifx #1\expandafter \@firstoftwo
 \else \expandafter \@secondoftwo
 \fi
}%
\providecommand \natexlab [1]{#1}%
\providecommand \enquote  [1]{``#1''}%
\providecommand \bibnamefont  [1]{#1}%
\providecommand \bibfnamefont [1]{#1}%
\providecommand \citenamefont [1]{#1}%
\providecommand \href@noop [0]{\@secondoftwo}%
\providecommand \href [0]{\begingroup \@sanitize@url \@href}%
\providecommand \@href[1]{\@@startlink{#1}\@@href}%
\providecommand \@@href[1]{\endgroup#1\@@endlink}%
\providecommand \@sanitize@url [0]{\catcode `\\12\catcode `\$12\catcode
  `\&12\catcode `\#12\catcode `\^12\catcode `\_12\catcode `\%12\relax}%
\providecommand \@@startlink[1]{}%
\providecommand \@@endlink[0]{}%
\providecommand \url  [0]{\begingroup\@sanitize@url \@url }%
\providecommand \@url [1]{\endgroup\@href {#1}{\urlprefix }}%
\providecommand \urlprefix  [0]{URL }%
\providecommand \Eprint [0]{\href }%
\providecommand \doibase [0]{http://dx.doi.org/}%
\providecommand \selectlanguage [0]{\@gobble}%
\providecommand \bibinfo  [0]{\@secondoftwo}%
\providecommand \bibfield  [0]{\@secondoftwo}%
\providecommand \translation [1]{[#1]}%
\providecommand \BibitemOpen [0]{}%
\providecommand \bibitemStop [0]{}%
\providecommand \bibitemNoStop [0]{.\EOS\space}%
\providecommand \EOS [0]{\spacefactor3000\relax}%
\providecommand \BibitemShut  [1]{\csname bibitem#1\endcsname}%
\let\auto@bib@innerbib\@empty
\bibitem [{\citenamefont {Workman}\ and\ \citenamefont
  {Others}(2022)}]{PDG2022}%
  \BibitemOpen
  \bibfield  {author} {\bibinfo {author} {\bibfnamefont {R.~L.}\ \bibnamefont
  {Workman}}\ and\ \bibinfo {author} {\bibnamefont {Others}} (\bibinfo
  {collaboration} {Particle Data Group}),\ }\href {\doibase
  10.1093/ptep/ptac097} {\bibfield  {journal} {\bibinfo  {journal} {PTEP}\
  }\textbf {\bibinfo {volume} {2022}},\ \bibinfo {pages} {083C01} (\bibinfo
  {year} {2022})}\BibitemShut {NoStop}%
\bibitem [{\citenamefont {de~Souza}\ \emph {et~al.}(2023)\citenamefont
  {de~Souza}, \citenamefont {Rom{\~a}o}, \citenamefont {Castro}, \citenamefont
  {Nikjoo},\ and\ \citenamefont {Porod}}]{de2023exploring}%
  \BibitemOpen
  \bibfield  {author} {\bibinfo {author} {\bibfnamefont {F.~A.}\ \bibnamefont
  {de~Souza}}, \bibinfo {author} {\bibfnamefont {M.~C.}\ \bibnamefont
  {Rom{\~a}o}}, \bibinfo {author} {\bibfnamefont {N.~F.}\ \bibnamefont
  {Castro}}, \bibinfo {author} {\bibfnamefont {M.}~\bibnamefont {Nikjoo}}, \
  and\ \bibinfo {author} {\bibfnamefont {W.}~\bibnamefont {Porod}},\
  }\href@noop {} {\bibfield  {journal} {\bibinfo  {journal} {Physical Review
  D}\ }\textbf {\bibinfo {volume} {107}},\ \bibinfo {pages} {035004} (\bibinfo
  {year} {2023})}\BibitemShut {NoStop}%
\bibitem [{\citenamefont {Holland}(1992)}]{holland1992adaptation}%
  \BibitemOpen
  \bibfield  {author} {\bibinfo {author} {\bibfnamefont {J.~H.}\ \bibnamefont
  {Holland}},\ }\href@noop {} {\emph {\bibinfo {title} {Adaptation in natural
  and artificial systems: an introductory analysis with applications to
  biology, control, and artificial intelligence}}}\ (\bibinfo  {publisher} {MIT
  press},\ \bibinfo {year} {1992})\BibitemShut {NoStop}%
\bibitem [{\citenamefont {Shapiro}(1999)}]{shapiro1999genetic}%
  \BibitemOpen
  \bibfield  {author} {\bibinfo {author} {\bibfnamefont {J.}~\bibnamefont
  {Shapiro}},\ }in\ \href@noop {} {\emph {\bibinfo {booktitle} {Advanced Course
  on Artificial Intelligence}}}\ (\bibinfo  {publisher} {Springer},\ \bibinfo
  {year} {1999})\ pp.\ \bibinfo {pages} {146--168}\BibitemShut {NoStop}%
\bibitem [{\citenamefont {Goldberg}\ and\ \citenamefont
  {Sastry}(2007)}]{goldberg2007genetic}%
  \BibitemOpen
  \bibfield  {author} {\bibinfo {author} {\bibfnamefont {D.}~\bibnamefont
  {Goldberg}}\ and\ \bibinfo {author} {\bibfnamefont {K.}~\bibnamefont
  {Sastry}},\ }\href@noop {} {\emph {\bibinfo {title} {Genetic algorithms: the
  design of innovation}}}\ (\bibinfo  {publisher} {Springer},\ \bibinfo {year}
  {2007})\BibitemShut {NoStop}%
\bibitem [{\citenamefont {Vellasco}\ \emph {et~al.}(2001)\citenamefont
  {Vellasco}, \citenamefont {Zebulum},\ and\ \citenamefont
  {Pacheco}}]{Vellasco:2001:EEA:559960}%
  \BibitemOpen
  \bibfield  {author} {\bibinfo {author} {\bibfnamefont {M.~M.~B.}\
  \bibnamefont {Vellasco}}, \bibinfo {author} {\bibfnamefont {R.~S.}\
  \bibnamefont {Zebulum}}, \ and\ \bibinfo {author} {\bibfnamefont {M.~A.}\
  \bibnamefont {Pacheco}},\ }\href@noop {} {\emph {\bibinfo {title}
  {Evolutionary Electronics: Automatic Design of Electronic Circuits and
  Systems by Genetic Algorithms}}},\ \bibinfo {edition} {1st}\ ed.\ (\bibinfo
  {publisher} {CRC Press, Inc.},\ \bibinfo {address} {Boca Raton, FL, USA},\
  \bibinfo {year} {2001})\BibitemShut {NoStop}%
\bibitem [{\citenamefont {Stanislawska}\ \emph {et~al.}(2012)\citenamefont
  {Stanislawska}, \citenamefont {Krawiec},\ and\ \citenamefont
  {Kundzewicz}}]{Stanislawska:2012:MGT:2400749.2401077}%
  \BibitemOpen
  \bibfield  {author} {\bibinfo {author} {\bibfnamefont {K.}~\bibnamefont
  {Stanislawska}}, \bibinfo {author} {\bibfnamefont {K.}~\bibnamefont
  {Krawiec}}, \ and\ \bibinfo {author} {\bibfnamefont {Z.~W.}\ \bibnamefont
  {Kundzewicz}},\ }\href {\doibase 10.1016/j.camwa.2012.02.049} {\bibfield
  {journal} {\bibinfo  {journal} {Comput. Math. Appl.}\ }\textbf {\bibinfo
  {volume} {64}},\ \bibinfo {pages} {3717} (\bibinfo {year}
  {2012})}\BibitemShut {NoStop}%
\bibitem [{\citenamefont {Sugawara}(2001)}]{Sugawara2001}%
  \BibitemOpen
  \bibfield  {author} {\bibinfo {author} {\bibfnamefont {M.}~\bibnamefont
  {Sugawara}},\ }\href@noop {} {\bibfield  {journal} {\bibinfo  {journal}
  {Computer Physics Communications}\ }\textbf {\bibinfo {volume} {140}},\
  \bibinfo {pages} {366} (\bibinfo {year} {2001})}\BibitemShut {NoStop}%
\bibitem [{\citenamefont {Guo}\ \emph {et~al.}(2023)\citenamefont {Guo},
  \citenamefont {Wang}, \citenamefont {Guo}, \citenamefont {Li}, \citenamefont
  {Song}, \citenamefont {Tan}, \citenamefont {Liu}, \citenamefont {Bian},\ and\
  \citenamefont {Yang}}]{Guo2023}%
  \BibitemOpen
  \bibfield  {author} {\bibinfo {author} {\bibfnamefont {Q.}~\bibnamefont
  {Guo}}, \bibinfo {author} {\bibfnamefont {R.}~\bibnamefont {Wang}}, \bibinfo
  {author} {\bibfnamefont {J.}~\bibnamefont {Guo}}, \bibinfo {author}
  {\bibfnamefont {B.}~\bibnamefont {Li}}, \bibinfo {author} {\bibfnamefont
  {K.}~\bibnamefont {Song}}, \bibinfo {author} {\bibfnamefont {X.}~\bibnamefont
  {Tan}}, \bibinfo {author} {\bibfnamefont {G.}~\bibnamefont {Liu}}, \bibinfo
  {author} {\bibfnamefont {J.}~\bibnamefont {Bian}}, \ and\ \bibinfo {author}
  {\bibfnamefont {Y.}~\bibnamefont {Yang}},\ }\href
  {http://arxiv.org/abs/2309.08532} {\  (\bibinfo {year} {2023})}\BibitemShut
  {NoStop}%
\bibitem [{\citenamefont {Unger}\ and\ \citenamefont
  {Moult}(1993)}]{Unger1993}%
  \BibitemOpen
  \bibfield  {author} {\bibinfo {author} {\bibfnamefont {R.}~\bibnamefont
  {Unger}}\ and\ \bibinfo {author} {\bibfnamefont {J.}~\bibnamefont {Moult}},\
  }\href {\doibase 10.1006/JMBI.1993.1258} {\bibfield  {journal} {\bibinfo
  {journal} {Journal of Molecular Biology}\ }\textbf {\bibinfo {volume}
  {231}},\ \bibinfo {pages} {75} (\bibinfo {year} {1993})}\BibitemShut
  {NoStop}%
\bibitem [{\citenamefont {Pedersen}\ and\ \citenamefont
  {Moult}(1997)}]{Pedersen1997}%
  \BibitemOpen
  \bibfield  {author} {\bibinfo {author} {\bibfnamefont {J.~T.}\ \bibnamefont
  {Pedersen}}\ and\ \bibinfo {author} {\bibfnamefont {J.}~\bibnamefont
  {Moult}},\ }\href@noop {} {\bibfield  {journal} {\bibinfo  {journal} {Journal
  of Molecular Biology}\ }\textbf {\bibinfo {volume} {269}},\ \bibinfo {pages}
  {240} (\bibinfo {year} {1997})}\BibitemShut {NoStop}%
\bibitem [{\citenamefont {Chew}\ \emph {et~al.}(2023)\citenamefont {Chew},
  \citenamefont {Joseph}, \citenamefont {Collepardo-Guevara},\ and\
  \citenamefont {Reinhardt}}]{Chew2023}%
  \BibitemOpen
  \bibfield  {author} {\bibinfo {author} {\bibfnamefont {P.~Y.}\ \bibnamefont
  {Chew}}, \bibinfo {author} {\bibfnamefont {J.~A.}\ \bibnamefont {Joseph}},
  \bibinfo {author} {\bibfnamefont {R.}~\bibnamefont {Collepardo-Guevara}}, \
  and\ \bibinfo {author} {\bibfnamefont {A.}~\bibnamefont {Reinhardt}},\ }\href
  {\doibase 10.1039/D2SC05873A} {\bibfield  {journal} {\bibinfo  {journal}
  {Chemical Science}\ }\textbf {\bibinfo {volume} {14}},\ \bibinfo {pages}
  {1820} (\bibinfo {year} {2023})}\BibitemShut {NoStop}%
\bibitem [{\citenamefont {Allanach}\ \emph {et~al.}(2004)\citenamefont
  {Allanach}, \citenamefont {Grellscheid},\ and\ \citenamefont
  {Quevedo}}]{Allanach2004}%
  \BibitemOpen
  \bibfield  {author} {\bibinfo {author} {\bibfnamefont {B.~C.}\ \bibnamefont
  {Allanach}}, \bibinfo {author} {\bibfnamefont {D.}~\bibnamefont
  {Grellscheid}}, \ and\ \bibinfo {author} {\bibfnamefont {F.}~\bibnamefont
  {Quevedo}},\ }\href {\doibase 10.1088/1126-6708/2004/07/069} {\bibfield
  {journal} {\bibinfo  {journal} {Journal of High Energy Physics}\ }\textbf
  {\bibinfo {volume} {2004}},\ \bibinfo {pages} {069} (\bibinfo {year}
  {2004})}\BibitemShut {NoStop}%
\bibitem [{\citenamefont {Cranmer}\ and\ \citenamefont
  {Bowman}(2005)}]{Cranmer2005}%
  \BibitemOpen
  \bibfield  {author} {\bibinfo {author} {\bibfnamefont {K.}~\bibnamefont
  {Cranmer}}\ and\ \bibinfo {author} {\bibfnamefont {R.~S.}\ \bibnamefont
  {Bowman}},\ }\href {\doibase 10.1016/j.cpc.2004.12.006} {\bibfield  {journal}
  {\bibinfo  {journal} {Computer Physics Communications}\ }\textbf {\bibinfo
  {volume} {167}},\ \bibinfo {pages} {165} (\bibinfo {year}
  {2005})}\BibitemShut {NoStop}%
\bibitem [{\citenamefont {Teodorescu}\ and\ \citenamefont
  {Sherwood}(2008)}]{Teodorescu2008}%
  \BibitemOpen
  \bibfield  {author} {\bibinfo {author} {\bibfnamefont {L.}~\bibnamefont
  {Teodorescu}}\ and\ \bibinfo {author} {\bibfnamefont {D.}~\bibnamefont
  {Sherwood}},\ }\href {\doibase 10.1016/j.cpc.2007.10.003} {\bibfield
  {journal} {\bibinfo  {journal} {Computer Physics Communications}\ }\textbf
  {\bibinfo {volume} {178}},\ \bibinfo {pages} {409} (\bibinfo {year}
  {2008})}\BibitemShut {NoStop}%
\bibitem [{\citenamefont {Akrami}\ \emph {et~al.}(2010)\citenamefont {Akrami},
  \citenamefont {Scott}, \citenamefont {Edsj{\"o}}, \citenamefont {Conrad},\
  and\ \citenamefont {Bergstr{\"o}m}}]{Akrami2010}%
  \BibitemOpen
  \bibfield  {author} {\bibinfo {author} {\bibfnamefont {Y.}~\bibnamefont
  {Akrami}}, \bibinfo {author} {\bibfnamefont {P.}~\bibnamefont {Scott}},
  \bibinfo {author} {\bibfnamefont {J.}~\bibnamefont {Edsj{\"o}}}, \bibinfo
  {author} {\bibfnamefont {J.}~\bibnamefont {Conrad}}, \ and\ \bibinfo {author}
  {\bibfnamefont {L.}~\bibnamefont {Bergstr{\"o}m}},\ }\href {\doibase
  10.1007/JHEP04(2010)057} {\bibfield  {journal} {\bibinfo  {journal} {Journal
  of High Energy Physics}\ }\textbf {\bibinfo {volume} {2010}} (\bibinfo {year}
  {2010}),\ 10.1007/JHEP04(2010)057}\BibitemShut {NoStop}%
\bibitem [{\citenamefont {Ruehle}(2017)}]{Ruehle:2017mzq}%
  \BibitemOpen
  \bibfield  {author} {\bibinfo {author} {\bibfnamefont {F.}~\bibnamefont
  {Ruehle}},\ }\href {\doibase 10.1007/JHEP08(2017)038} {\bibfield  {journal}
  {\bibinfo  {journal} {JHEP}\ }\textbf {\bibinfo {volume} {08}},\ \bibinfo
  {pages} {038} (\bibinfo {year} {2017})},\ \Eprint
  {http://arxiv.org/abs/1706.07024} {arXiv:1706.07024 [hep-th]} \BibitemShut
  {NoStop}%
\bibitem [{\citenamefont {Camargo-Molina}\ \emph {et~al.}(2018)\citenamefont
  {Camargo-Molina}, \citenamefont {Mandal}, \citenamefont {Pasechnik},\ and\
  \citenamefont {Wess\'en}}]{Camargo-Molina:2017klw}%
  \BibitemOpen
  \bibfield  {author} {\bibinfo {author} {\bibfnamefont {J.~E.}\ \bibnamefont
  {Camargo-Molina}}, \bibinfo {author} {\bibfnamefont {T.}~\bibnamefont
  {Mandal}}, \bibinfo {author} {\bibfnamefont {R.}~\bibnamefont {Pasechnik}}, \
  and\ \bibinfo {author} {\bibfnamefont {J.}~\bibnamefont {Wess\'en}},\ }\href
  {\doibase 10.1007/JHEP03(2018)024} {\bibfield  {journal} {\bibinfo  {journal}
  {JHEP}\ }\textbf {\bibinfo {volume} {03}},\ \bibinfo {pages} {024} (\bibinfo
  {year} {2018})},\ \Eprint {http://arxiv.org/abs/1711.03551} {arXiv:1711.03551
  [hep-ph]} \BibitemShut {NoStop}%
\bibitem [{\citenamefont {Luo}\ \emph {et~al.}(2020)\citenamefont {Luo},
  \citenamefont {Feng},\ and\ \citenamefont {Zhang}}]{luo2020genetic}%
  \BibitemOpen
  \bibfield  {author} {\bibinfo {author} {\bibfnamefont {X.-L.}\ \bibnamefont
  {Luo}}, \bibinfo {author} {\bibfnamefont {J.}~\bibnamefont {Feng}}, \ and\
  \bibinfo {author} {\bibfnamefont {H.-H.}\ \bibnamefont {Zhang}},\ }\href@noop
  {} {\bibfield  {journal} {\bibinfo  {journal} {Computer Physics
  Communications}\ }\textbf {\bibinfo {volume} {250}},\ \bibinfo {pages}
  {106818} (\bibinfo {year} {2020})}\BibitemShut {NoStop}%
\bibitem [{\citenamefont {Biek{\"o}tter}\ and\ \citenamefont
  {Olea-Romacho}(2021)}]{biekotter2021reconciling}%
  \BibitemOpen
  \bibfield  {author} {\bibinfo {author} {\bibfnamefont {T.}~\bibnamefont
  {Biek{\"o}tter}}\ and\ \bibinfo {author} {\bibfnamefont {M.~O.}\ \bibnamefont
  {Olea-Romacho}},\ }\href@noop {} {\bibfield  {journal} {\bibinfo  {journal}
  {Journal of High Energy Physics}\ }\textbf {\bibinfo {volume} {2021}},\
  \bibinfo {pages} {1} (\bibinfo {year} {2021})}\BibitemShut {NoStop}%
\bibitem [{\citenamefont {Pal}\ \emph {et~al.}(2024)\citenamefont {Pal},
  \citenamefont {Wess\'en}, \citenamefont {Das},\ and\ \citenamefont
  {Chan}}]{Pal2024}%
  \BibitemOpen
  \bibfield  {author} {\bibinfo {author} {\bibfnamefont {T.}~\bibnamefont
  {Pal}}, \bibinfo {author} {\bibfnamefont {J.}~\bibnamefont {Wess\'en}},
  \bibinfo {author} {\bibfnamefont {S.}~\bibnamefont {Das}}, \ and\ \bibinfo
  {author} {\bibfnamefont {H.~S.}\ \bibnamefont {Chan}},\ }\href {\doibase
  10.1021/acs.jpclett.4c01973} {\bibfield  {journal} {\bibinfo  {journal} {The
  Journal of Physical Chemistry Letters}\ }\textbf {\bibinfo {volume} {15}},\
  \bibinfo {pages} {8248} (\bibinfo {year} {2024})},\ \bibinfo {note} {pMID:
  39105804}\BibitemShut {NoStop}%
\bibitem [{\citenamefont {Das}\ and\ \citenamefont {Pappu}(2013)}]{Das2013}%
  \BibitemOpen
  \bibfield  {author} {\bibinfo {author} {\bibfnamefont {R.~K.}\ \bibnamefont
  {Das}}\ and\ \bibinfo {author} {\bibfnamefont {R.~V.}\ \bibnamefont
  {Pappu}},\ }\href {\doibase 10.1073/PNAS.1304749110} {\bibfield  {journal}
  {\bibinfo  {journal} {Proceedings of the National Academy of Sciences of the
  United States of America}\ }\textbf {\bibinfo {volume} {110}},\ \bibinfo
  {pages} {13392} (\bibinfo {year} {2013})}\BibitemShut {NoStop}%
\bibitem [{\citenamefont {Sawle}\ and\ \citenamefont
  {Ghosh}(2015)}]{Sawle2015}%
  \BibitemOpen
  \bibfield  {author} {\bibinfo {author} {\bibfnamefont {L.}~\bibnamefont
  {Sawle}}\ and\ \bibinfo {author} {\bibfnamefont {K.}~\bibnamefont {Ghosh}},\
  }\href {\doibase 10.1063/1.4929391} {\bibfield  {journal} {\bibinfo
  {journal} {The Journal of Chemical Physics}\ }\textbf {\bibinfo {volume}
  {143}},\ \bibinfo {pages} {085101} (\bibinfo {year} {2015})}\BibitemShut
  {NoStop}%
\bibitem [{\citenamefont {Lin}\ \emph {et~al.}(2016)\citenamefont {Lin},
  \citenamefont {Forman-Kay},\ and\ \citenamefont {Chan}}]{Lin2016}%
  \BibitemOpen
  \bibfield  {author} {\bibinfo {author} {\bibfnamefont {Y.~H.}\ \bibnamefont
  {Lin}}, \bibinfo {author} {\bibfnamefont {J.~D.}\ \bibnamefont {Forman-Kay}},
  \ and\ \bibinfo {author} {\bibfnamefont {H.~S.}\ \bibnamefont {Chan}},\
  }\href {\doibase 10.1103/PhysRevLett.117.178101} {\bibfield  {journal}
  {\bibinfo  {journal} {Physical Review Letters}\ }\textbf {\bibinfo {volume}
  {117}} (\bibinfo {year} {2016}),\ 10.1103/PhysRevLett.117.178101}\BibitemShut
  {NoStop}%
\bibitem [{\citenamefont {McCarty}\ \emph {et~al.}(2019)\citenamefont
  {McCarty}, \citenamefont {Delaney}, \citenamefont {Danielsen}, \citenamefont
  {Fredrickson},\ and\ \citenamefont {Shea}}]{McCarty2019}%
  \BibitemOpen
  \bibfield  {author} {\bibinfo {author} {\bibfnamefont {J.}~\bibnamefont
  {McCarty}}, \bibinfo {author} {\bibfnamefont {K.~T.}\ \bibnamefont
  {Delaney}}, \bibinfo {author} {\bibfnamefont {S.~P.}\ \bibnamefont
  {Danielsen}}, \bibinfo {author} {\bibfnamefont {G.~H.}\ \bibnamefont
  {Fredrickson}}, \ and\ \bibinfo {author} {\bibfnamefont {J.~E.}\ \bibnamefont
  {Shea}},\ }\href {\doibase 10.1021/acs.jpclett.9b00099} {\bibfield  {journal}
  {\bibinfo  {journal} {Journal of Physical Chemistry Letters}\ }\textbf
  {\bibinfo {volume} {10}},\ \bibinfo {pages} {1644} (\bibinfo {year}
  {2019})}\BibitemShut {NoStop}%
\bibitem [{\citenamefont {Pal}\ \emph {et~al.}(2021)\citenamefont {Pal},
  \citenamefont {Wess\'en}, \citenamefont {Das},\ and\ \citenamefont
  {Chan}}]{Pal2021}%
  \BibitemOpen
  \bibfield  {author} {\bibinfo {author} {\bibfnamefont {T.}~\bibnamefont
  {Pal}}, \bibinfo {author} {\bibfnamefont {J.}~\bibnamefont {Wess\'en}},
  \bibinfo {author} {\bibfnamefont {S.}~\bibnamefont {Das}}, \ and\ \bibinfo
  {author} {\bibfnamefont {H.~S.}\ \bibnamefont {Chan}},\ }\href {\doibase
  10.1103/PhysRevE.103.042406} {\bibfield  {journal} {\bibinfo  {journal}
  {Physical Review E}\ }\textbf {\bibinfo {volume} {103}} (\bibinfo {year}
  {2021}),\ 10.1103/PhysRevE.103.042406}\BibitemShut {NoStop}%
\bibitem [{\citenamefont {Wess\'en}\ \emph {et~al.}(2021)\citenamefont
  {Wess\'en}, \citenamefont {Pal}, \citenamefont {Das}, \citenamefont {Lin},\
  and\ \citenamefont {Chan}}]{Wessen2021}%
  \BibitemOpen
  \bibfield  {author} {\bibinfo {author} {\bibfnamefont {J.}~\bibnamefont
  {Wess\'en}}, \bibinfo {author} {\bibfnamefont {T.}~\bibnamefont {Pal}},
  \bibinfo {author} {\bibfnamefont {S.}~\bibnamefont {Das}}, \bibinfo {author}
  {\bibfnamefont {Y.~H.}\ \bibnamefont {Lin}}, \ and\ \bibinfo {author}
  {\bibfnamefont {H.~S.}\ \bibnamefont {Chan}},\ }\href {\doibase
  10.1021/acs.jpcb.1c00954} {\bibfield  {journal} {\bibinfo  {journal} {Journal
  of Physical Chemistry B}\ }\textbf {\bibinfo {volume} {125}},\ \bibinfo
  {pages} {4337} (\bibinfo {year} {2021})}\BibitemShut {NoStop}%
\bibitem [{\citenamefont {Wess\'en}\ \emph
  {et~al.}(2022{\natexlab{a}})\citenamefont {Wess\'en}, \citenamefont {Pal},\
  and\ \citenamefont {Chan}}]{Wessen2022JCP}%
  \BibitemOpen
  \bibfield  {author} {\bibinfo {author} {\bibfnamefont {J.}~\bibnamefont
  {Wess\'en}}, \bibinfo {author} {\bibfnamefont {T.}~\bibnamefont {Pal}}, \
  and\ \bibinfo {author} {\bibfnamefont {H.~S.}\ \bibnamefont {Chan}},\ }\href
  {\doibase 10.1063/5.0088326} {\bibfield  {journal} {\bibinfo  {journal}
  {Journal of Chemical Physics}\ }\textbf {\bibinfo {volume} {156}} (\bibinfo
  {year} {2022}{\natexlab{a}}),\ 10.1063/5.0088326}\BibitemShut {NoStop}%
\bibitem [{\citenamefont {Wess\'en}\ \emph
  {et~al.}(2022{\natexlab{b}})\citenamefont {Wess\'en}, \citenamefont {Das},
  \citenamefont {Pal},\ and\ \citenamefont {Chan}}]{Wessen2022JPCB}%
  \BibitemOpen
  \bibfield  {author} {\bibinfo {author} {\bibfnamefont {J.}~\bibnamefont
  {Wess\'en}}, \bibinfo {author} {\bibfnamefont {S.}~\bibnamefont {Das}},
  \bibinfo {author} {\bibfnamefont {T.}~\bibnamefont {Pal}}, \ and\ \bibinfo
  {author} {\bibfnamefont {H.~S.}\ \bibnamefont {Chan}},\ }\href {\doibase
  10.1021/acs.jpcb.2c06181} {\bibfield  {journal} {\bibinfo  {journal} {The
  Journal of Physical Chemistry B}\ }\textbf {\bibinfo {volume} {126}},\
  \bibinfo {pages} {9222 } (\bibinfo {year} {2022}{\natexlab{b}})}\BibitemShut
  {NoStop}%
\bibitem [{\citenamefont {Lin}\ \emph {et~al.}(2023)\citenamefont {Lin},
  \citenamefont {Wess\'en}, \citenamefont {Pal}, \citenamefont {Das},\ and\
  \citenamefont {Chan}}]{Lin2023}%
  \BibitemOpen
  \bibfield  {author} {\bibinfo {author} {\bibfnamefont {Y.~H.}\ \bibnamefont
  {Lin}}, \bibinfo {author} {\bibfnamefont {J.}~\bibnamefont {Wess\'en}},
  \bibinfo {author} {\bibfnamefont {T.}~\bibnamefont {Pal}}, \bibinfo {author}
  {\bibfnamefont {S.}~\bibnamefont {Das}}, \ and\ \bibinfo {author}
  {\bibfnamefont {H.~S.}\ \bibnamefont {Chan}},\ }\href {\doibase
  10.1007/978-1-0716-2663-4{\_}3} {\bibfield  {journal} {\bibinfo  {journal}
  {Methods in Molecular Biology}\ }\textbf {\bibinfo {volume} {2563}} (\bibinfo
  {year} {2023}),\ 10.1007/978-1-0716-2663-4{\_}3}\BibitemShut {NoStop}%
\bibitem [{\citenamefont {{Blank}}\ and\ \citenamefont {{Deb}}(2020)}]{pymoo}%
  \BibitemOpen
  \bibfield  {author} {\bibinfo {author} {\bibfnamefont {J.}~\bibnamefont
  {{Blank}}}\ and\ \bibinfo {author} {\bibfnamefont {K.}~\bibnamefont
  {{Deb}}},\ }\href@noop {} {\bibfield  {journal} {\bibinfo  {journal} {IEEE
  Access}\ }\textbf {\bibinfo {volume} {8}},\ \bibinfo {pages} {89497}
  (\bibinfo {year} {2020})}\BibitemShut {NoStop}%
\bibitem [{\citenamefont {Gad}(2021)}]{gad2021pygad}%
  \BibitemOpen
  \bibfield  {author} {\bibinfo {author} {\bibfnamefont {A.~F.}\ \bibnamefont
  {Gad}},\ }\href@noop {} {\enquote {\bibinfo {title} {Pygad: An intuitive
  genetic algorithm python library},}\ } (\bibinfo {year} {2021}),\ \Eprint
  {http://arxiv.org/abs/2106.06158} {arXiv:2106.06158 [cs.NE]} \BibitemShut
  {NoStop}%
\bibitem [{\citenamefont {Branco}\ \emph {et~al.}(2012)\citenamefont {Branco},
  \citenamefont {Ferreira}, \citenamefont {Lavoura}, \citenamefont {Rebelo},
  \citenamefont {Sher} \emph {et~al.}}]{Branco:2011iw}%
  \BibitemOpen
  \bibfield  {author} {\bibinfo {author} {\bibfnamefont {G.}~\bibnamefont
  {Branco}}, \bibinfo {author} {\bibfnamefont {P.}~\bibnamefont {Ferreira}},
  \bibinfo {author} {\bibfnamefont {L.}~\bibnamefont {Lavoura}}, \bibinfo
  {author} {\bibfnamefont {M.}~\bibnamefont {Rebelo}}, \bibinfo {author}
  {\bibfnamefont {M.}~\bibnamefont {Sher}},  \emph {et~al.},\ }\href {\doibase
  10.1016/j.physrep.2012.02.002} {\bibfield  {journal} {\bibinfo  {journal}
  {Phys.Rept.}\ }\textbf {\bibinfo {volume} {516}},\ \bibinfo {pages} {1}
  (\bibinfo {year} {2012})},\ \Eprint {http://arxiv.org/abs/1106.0034}
  {arXiv:1106.0034 [hep-ph]} \BibitemShut {NoStop}%
\bibitem [{\citenamefont {Bechtle}\ \emph
  {et~al.}(2014{\natexlab{a}})\citenamefont {Bechtle}, \citenamefont {Brein},
  \citenamefont {Heinemeyer}, \citenamefont {St\r{a}l}, \citenamefont
  {Stefaniak}, \citenamefont {Weiglein},\ and\ \citenamefont
  {Williams}}]{HiggsBounds}%
  \BibitemOpen
  \bibfield  {author} {\bibinfo {author} {\bibfnamefont {P.}~\bibnamefont
  {Bechtle}}, \bibinfo {author} {\bibfnamefont {O.}~\bibnamefont {Brein}},
  \bibinfo {author} {\bibfnamefont {S.}~\bibnamefont {Heinemeyer}}, \bibinfo
  {author} {\bibfnamefont {O.}~\bibnamefont {St\r{a}l}}, \bibinfo {author}
  {\bibfnamefont {T.}~\bibnamefont {Stefaniak}}, \bibinfo {author}
  {\bibfnamefont {G.}~\bibnamefont {Weiglein}}, \ and\ \bibinfo {author}
  {\bibfnamefont {K.~E.}\ \bibnamefont {Williams}},\ }\href {\doibase
  10.1140/epjc/s10052-013-2693-2} {\bibfield  {journal} {\bibinfo  {journal}
  {Eur. Phys. J. C}\ }\textbf {\bibinfo {volume} {74}},\ \bibinfo {pages}
  {2693} (\bibinfo {year} {2014}{\natexlab{a}})},\ \Eprint
  {http://arxiv.org/abs/1311.0055} {arXiv:1311.0055 [hep-ph]} \BibitemShut
  {NoStop}%
\bibitem [{\citenamefont {Eriksson}\ \emph {et~al.}(2010)\citenamefont
  {Eriksson}, \citenamefont {Rathsman},\ and\ \citenamefont
  {St\r{a}l}}]{2HDMC}%
  \BibitemOpen
  \bibfield  {author} {\bibinfo {author} {\bibfnamefont {D.}~\bibnamefont
  {Eriksson}}, \bibinfo {author} {\bibfnamefont {J.}~\bibnamefont {Rathsman}},
  \ and\ \bibinfo {author} {\bibfnamefont {O.}~\bibnamefont {St\r{a}l}},\
  }\href {\doibase 10.1016/j.cpc.2009.09.011} {\bibfield  {journal} {\bibinfo
  {journal} {Comput. Phys. Commun.}\ }\textbf {\bibinfo {volume} {181}},\
  \bibinfo {pages} {189} (\bibinfo {year} {2010})},\ \Eprint
  {http://arxiv.org/abs/0902.0851} {arXiv:0902.0851 [hep-ph]} \BibitemShut
  {NoStop}%
\bibitem [{\citenamefont {Bechtle}\ \emph
  {et~al.}(2014{\natexlab{b}})\citenamefont {Bechtle}, \citenamefont
  {Heinemeyer}, \citenamefont {St\r{a}l}, \citenamefont {Stefaniak},\ and\
  \citenamefont {Weiglein}}]{HiggsSignals}%
  \BibitemOpen
  \bibfield  {author} {\bibinfo {author} {\bibfnamefont {P.}~\bibnamefont
  {Bechtle}}, \bibinfo {author} {\bibfnamefont {S.}~\bibnamefont {Heinemeyer}},
  \bibinfo {author} {\bibfnamefont {O.}~\bibnamefont {St\r{a}l}}, \bibinfo
  {author} {\bibfnamefont {T.}~\bibnamefont {Stefaniak}}, \ and\ \bibinfo
  {author} {\bibfnamefont {G.}~\bibnamefont {Weiglein}},\ }\href {\doibase
  10.1140/epjc/s10052-013-2711-4} {\bibfield  {journal} {\bibinfo  {journal}
  {Eur. Phys. J. C}\ }\textbf {\bibinfo {volume} {74}},\ \bibinfo {pages}
  {2711} (\bibinfo {year} {2014}{\natexlab{b}})},\ \Eprint
  {http://arxiv.org/abs/1305.1933} {arXiv:1305.1933 [hep-ph]} \BibitemShut
  {NoStop}%
\bibitem [{\citenamefont {Das}\ and\ \citenamefont
  {Suganthan}(2011)}]{5601760}%
  \BibitemOpen
  \bibfield  {author} {\bibinfo {author} {\bibfnamefont {S.}~\bibnamefont
  {Das}}\ and\ \bibinfo {author} {\bibfnamefont {P.~N.}\ \bibnamefont
  {Suganthan}},\ }\href {\doibase 10.1109/TEVC.2010.2059031} {\bibfield
  {journal} {\bibinfo  {journal} {IEEE Transactions on Evolutionary
  Computation}\ }\textbf {\bibinfo {volume} {15}},\ \bibinfo {pages} {4}
  (\bibinfo {year} {2011})}\BibitemShut {NoStop}%
\end{thebibliography}%
\appendix
\counterwithin{figure}{section}
\renewcommand{\thefigure}{A\arabic{figure}}
\section{Lightweight Genetic Algorithm Package} \label{sec:lga_documentation}

\subsection{Installation}
The GA presented in this work is implemented in the \texttt{lightweight-genetic-algorithm} Python module which is installable via pip as
\begin{lstlisting}[language=bash]
pip install lightweight-genetic-algorithm
\end{lstlisting}
The source code and documentation can be found at:
\begin{center}
\href{https://github.com/JoseEliel/lightweight_genetic_algorithm}{github.com/JoseEliel/lightweight\_genetic\_algorithm} (Source Code)\\
\href{https://lightweight-genetic-algorithm.readthedocs.io/}{lightweight-genetic-algorithm.readthedocs.io} (Documentation)
\end{center}

\subsection{Features}

The \texttt{lightweight-genetic-algorithm} Python module contains several features that allows the user to easily set up a GA for a wide range types of optimization problems. These features include

\begin{itemize}
    \item \textbf{Support for Numerical and Categorical Genes}: The package can handle optimisation problems formulated either in terms of numerical- or categorical genes.

    \item \textbf{Multiple Crossover Methods}: The package provides four different crossover methods: \textit{Between}, \textit{Midpoint}, \textit{Either Or} and \textit{None}. These methods are described in Sec.~\ref{sec:crossover_methods}.

    \item \textbf{Multiple Mutation Modes}: The package includes three mutation modes for numerical genes: \textit{additive}, \textit{multiplicative}, \textit{random}, described in Sec.~\ref{sec:crossover_methods}. For categorical genes, the algorithm assumes the \textit{categorical} mutation mode described in Sec.~\ref{sec:adaptations_categorical}.

    \item \textbf{Diversity-Enhanced Selection}: The package uses the diversity-enhanced selection algorithm, described in Sec.~\ref{sec:diversity_enhanced_selection}, through which the algorithm can simultaneously explore widely different regions in parameter space. 

    \item \textbf{Customizable Distance Measure}: The user can specify the distance function (used during selection) to be either \textit{Euclidean} or \textit{Dynamic}, defined respectively in Eqs.~\eqref{eq:euclidean_distance_measure} and \eqref{eq:dynamic_distance_measure}, for numeric genes. For categorical genes, the Hamming distance in Eq.~\eqref{eq:hamming_distance_measure} is assumed. It is also possible for the user to supply their own distance function.
  
    \item \textbf{Multiprocessing}: The package supports multiprocessing for parallel fitness evaluations. This feature can dramatically speed up the genetic algorithm for problems where the fitness function is computationally expensive.
\end{itemize}

\subsection{User guide}

The primary class in this package is \texttt{GeneticAlgorithm}. A \texttt{GeneticAlgorithm} instance is created with the required input arguments:
\begin{itemize}
    \item \texttt{fitness\_function}: A function computing the fitness score of an individual. This function should receive an array of genes as its first input argument and return a single number. Additional arguments can be passed to the fitness function using the \texttt{fitness\_function\_args} argument, described below.

    \item \texttt{gene\_ranges}: A list of tuples representing the range of each numeric gene. Each tuple should contain two numbers, with the first number being the lower bound and the second the upper bound. For categorical genes, \texttt{gene\_ranges} should instead be a one-dimensional list of possible categories.

    \item \texttt{number\_of\_genes} (only needed for categorical genes): The number of genes defining an individual. For numeric genes, the \texttt{number\_of\_genes} is inferred from the length of \texttt{gene\_ranges}.
\end{itemize}

Furthermore, the following optional arguments can be provided to the \texttt{GeneticAlgorithm} constructor: 

\begin{itemize}
    \item \texttt{fitness\_function\_args} (optional): Additional arguments to pass to the fitness function. This should be a tuple of arguments.

    \item \texttt{crossover\_method} (optional): The method used for crossover. Available options are \texttt{Between}, \texttt{Midpoint}, \texttt{Either Or} and \texttt{None}. Default is \texttt{Between} for numeric genes. For categorical genes, only \texttt{Either Or} or \texttt{None} is possible.

    \item \texttt{mutation\_mode} (optional): The mode used for mutation. Options available are \texttt{additive}, \texttt{multiplicative}, \texttt{random} and \texttt{categorical}. Default is \texttt{additive} for numeric genes and \texttt{categorical} for categorical genes.

    \item \texttt{mutation\_rate} (optional): The rate of mutation. The default is 1.0/\texttt{number\_of\_genes}. During crossover, each gene is mutated with probability \texttt{mutation\_rate}.

    \item \texttt{measure} (optional): Specifies the distance function between two points in the gene space. This argument can be a string variable (\texttt{Euclidean}, \texttt{Dynamic} or \texttt{Hamming}) corresponding to the three distance measures discussed in this work. The \texttt{measure} argument can also be a user-defined distance function. The default is Euclidean distance for numeric genes and Hamming distance for categorical genes.
    
    \item \texttt{r0} (optional): The characteristic distance beyond which there is no diversity penalty. Default is 1/10 of the average spread of initial population. Only used for diversity enhanced selection.

    \item \texttt{D0} (optional): The maximum diversity penalty for identical individuals. Default is 1.0. Only used for diversity enhanced selection.
    
    \item \texttt{use\_multiprocessing} (optional): Whether to use multiprocessing for parallel fitness evaluations. Default is False.

    \item \texttt{ncpus} (optional): The number of CPUs to use for multiprocessing. Default is the number of CPUs on the system minus one. This argument is used only when \texttt{use\_multiprocessing} is True.
    
    \item \texttt{selection\_method} (optional): The method used for survivor selection. Available options are \texttt{Diversity Enhanced} and \texttt{Fitness Proportionate}. Default is \texttt{Diversity Enhanced}. \texttt{Fitness Proportionate} simply selects the best individuals based on their fitness scores, while \texttt{Diversity Enhanced} uses the diversity-enhanced survivor selection method described above.
    
    \item \texttt{output\_directory} (optional): The directory where the output files are saved. Default is None. If specified, the algorithm saves the genes of the selected survivors in each generation to a file \texttt{<date-time>\_survivors.npy} in the output directory, where \texttt{<date-time>} is a timestamp indicating when the algorithm started. The average fitness and best fitness at each generation are saved to \texttt{<date-time>\_fitness.txt}. A log file is saved to \texttt{log.txt} containing the output of the algorithm printed to the console. The \texttt{<date-time>\_survivors.npy} file can be loaded using \texttt{numpy.load(file, allow\_pickle=True)} to access the gene values of the individuals in each generation. The loaded array has shape $(\mathtt{n\_generations}+1, \mathtt{population\_size}, \mathtt{number\_of\_genes})$ where the ``+1'' comes from the fact that the initial population is also saved.

\end{itemize}

Once an instance of the \texttt{GeneticAlgorithm} class has been created, the genetic algorithm is executed using the \texttt{run} or \texttt{run\_light} methods. These methods take the following arguments:

\begin{itemize}
    \item \texttt{n\_generations}: The number of generations to run the genetic algorithm for.

    \item \texttt{population\_size}: The number of individuals in the population.

    \item \texttt{fitness\_threshold} (optional): The fitness threshold at which the genetic algorithm should stop. If this is set, the genetic algorithm will stop when the fitness of the best individual in the population is greater than or equal to the fitness threshold. Default is None.

    \item \texttt{init\_genes} (optional): An initial set of gene values for creating the initial population. Default is \texttt{None}.

    \item \texttt{verbosity} (optional): The verbosity level for printing out messages. The options are \texttt{0} (silent), \texttt{1} (normal output) and \texttt{2} (detailed output). Default is \texttt{1}.
\end{itemize}

The \texttt{run} method returns the full list of \texttt{Individual} instances across all generations, where each \texttt{Individual} object has attributes such as fitness and gene values.

The \texttt{run\_light} method is similar to \texttt{run} but returns only the gene values of all individuals across all generations, which is useful for large populations and generations.

\subsection{Examples}
\subsubsection*{Example 1: Numerical genes}

Figure \ref{fig:numeric_code} contains the Python code for a simple example of how to use the package. In this example, an individual represents a point in the $xy$-plane, and the fitness function takes the form
\begin{equation}
    f(x,y) = - A \left( \sqrt{x^2 + y^2} - R \right)^2,
\end{equation}
which has an extended maximum on the circle centered at the origin with radius $R$. The overall factor $A$ can be chosen to balance the diversity punishment as scaling the over-all fitness by a factor $\xi$ is equivalent as scaling the diversity punishment parameter $D_0$ by $\xi^{-1}$, i.e.
\begin{equation}
    A \rightarrow \xi A \quad \Leftrightarrow \quad D_0 \rightarrow \xi^{-1} D_0.
\end{equation}

\begin{figure*}
    \centering
    \includegraphics{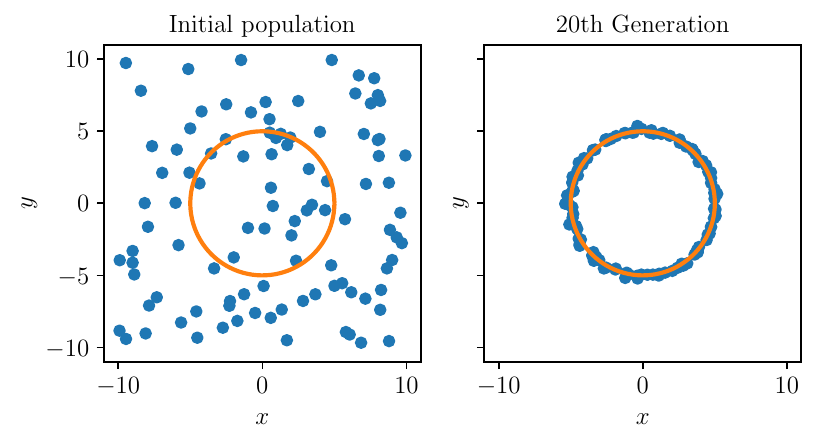}
    \caption{Initial and final populations for the numerical genes example. In just 20 generations, all the individuals are evenly distributed along the circle.}
    \label{fig:numerical_genes_example}
\end{figure*}

The resulting population after 20 generations is depicted in Figure \ref{fig:numerical_genes_example}. In 20 generations, the individuals are evenly distributed along the circle.

\begin{figure*}
    \begin{lstlisting}[language=Python]
from lightweight_genetic_algorithm import GeneticAlgorithm

# Define fitness function
def fitness_function(individual): # individual is x,y coordinates
    distance = (individual[0]**2 + individual[1]**2)**0.5
    R = 5 # Circle radius
    A = 5 # Overall fitness scaling 
    fitness = -A*(distance - R)**2 
    return fitness

# Define the ranges of the genes
gene_ranges = [ (-10,10), (-10,10) ]

# Create a GeneticAlgorithm instance
ga = GeneticAlgorithm(fitness_function, gene_ranges, crossover_method='Between')
all_populations = ga.run_light(n_generations=20, population_size=100)

# all_populations is a (n_generations, population_size, n_genes) list.
# The final population is all_populations[-1]
    \end{lstlisting}
    \caption{Complete Python code for the simple example for numeric genes. In this example, an individual represents coordinates of a point in the $xy$-plane. The fitness function has an extended maximum on a circle with radius 5.0. The genetic algorithm is run for 20 generations with a population size of 100 using the \texttt{Between} crossover method.}
    \label{fig:numeric_code}
\end{figure*}

\subsubsection*{Example 2: Categorical genes}

Next, we turn to a slightly more complex example involving categorical genes. In this example, we seek to construct an array of Lysine (K) Glutamic acid (E) representing the amino-acid sequence of a model intrinsically disordered protein. The goal is to find a diverse set of sequences with a sequence charge decoration (SCD) parameter near a given target value.

The net charge of a sequence is the sum of the charges of the amino acids with Lysine (K) having a charge of +1 and Glutamic Acid (E) having a charge of -1. The SCD parameter is defined in~\cite{Sawle2015}, and is a single number that can be calculated given a sequence of charges. The SCD parameter is a measure of the ``charge blockiness'' (i.e., an alternating sequence `EKEKEK \ldots EK' has $\mathrm{SCD} \approx 0$ while a di-block sequence `EEEE \ldots EEEKKKK \ldots KKK' gives a large, negative SCD) and correlates well with both the radius-of-gyration of isolated chains and with the upper-critical temperature for phase separation in multi-chain systems.

The complete Python code for this example is shown in Figure \ref{fig:categorical_code}. In this example, an individual corresponds to a list of \texttt{E}'s and \texttt{K}'s representing the amino-acid sequence. This code showcases two additional important features of the \texttt{lightweight-genetic-algorithm} module: multiprocessing and the usage of additional arguments to the fitness function. 

The multiprocessing is set up by providing the argument \texttt{use\_multiprocessing = True} to the \texttt{GeneticAlgorithm} constructor. Note that the main code is placed inside a \texttt{main()} function, which is run when the script is executed directly (i.e., not imported as a module). This is a requirement for the multiprocessing to work. The fitness function takes the sequence and the target SCD value \texttt{target\_SCD} as arguments, where \texttt{target\_SCD} is passed to the \texttt{GeneticAlgorithm} constructor as the \texttt{fitness\_function\_args} argument.

\begin{figure*}
    \centering
    \includegraphics[width=0.5\textwidth]{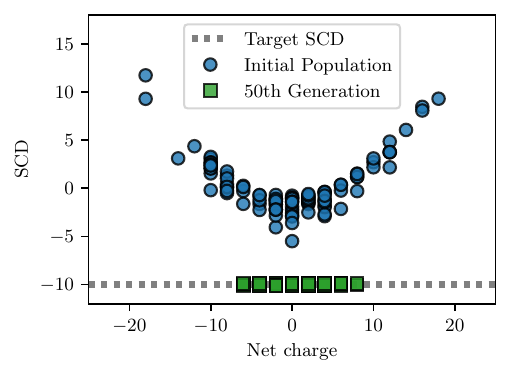}
    \caption{SCD values and net charges for initial and final populations of the categorical genes example.}
    \label{fig:categorical_genes_example}
\end{figure*}

The net charges and SCD values for the initial and final populations are shown in Figure \ref{fig:categorical_genes_example}. Note that the SCD values are close to the target value of -10 while there is a wide range of net charges in the final population. This demonstrates the effect of the diversity-enhanced selection method.

\begin{figure*}
    \begin{lstlisting}[language=Python]
from lightweight_genetic_algorithm import GeneticAlgorithm

# Calculates the sequence charge decoration (SCD) parameter
def calculate_SCD(sequence):
    aa_charges = {'K':1, 'E':-1} # Amino acid electric charges
    charge_sequence = [ aa_charges[aa] for aa in sequence ]
    SCD = 0
    for a in range(len(charge_sequence)-1):
        for b in range(a+1,len(charge_sequence)):
            SCD += charge_sequence[a] * charge_sequence[b] * abs(a-b)**0.5
    SCD /= len(charge_sequence)
    return SCD

# Define fitness function
def fitness_function(sequence, target_SCD):
    SCD = calculate_SCD(sequence)
    fitness = -(SCD - target_SCD)**2
    return fitness

def main():
    # Categorical genes are recognized automatically. 
    gene_ranges = ['E', 'K'] # Glutamic acid (E) , Lysine (K)
    N = 50 # Sequence length
    target_SCD = -10 # Target SCD value

    # Create a GeneticAlgorithm instance
    ga = GeneticAlgorithm(fitness_function, gene_ranges, 
                        number_of_genes = N, 
                        fitness_function_args = (target_SCD,), 
                        use_multiprocessing = True)

    # Run the genetic algorithm
    all_populations = ga.run_light(n_generations=50, population_size=100)

if __name__ == '__main__':
    main()
    \end{lstlisting}
    \caption{Complete Python code for the categorical genes example, showcasing the usage of multiprocessing and additional fitness function arguments. The GA is run for 50 generations with a population size of 100. The final population of sequences is contained in \texttt{all\_populations[-1]} which is a list of length 100 where each entry is a list of \texttt{E}'s and \texttt{K}'s representing the amino-acid sequence.}
    \label{fig:categorical_code}
\end{figure*}

\end{document}